# MotionTeller: Multi-modal Integration of Wearable Time-Series with LLMs for Health and Behavioral Understanding

Aiwei Zhang[1,2], Arvind Pillai[2], Andrew Campbell[2] & Nicholas C. Jacobson[1,2,3,4]

[1]Center for Technology and Behavioral Health, Geisel School of Medicine, Dartmouth College, Lebanon, NH, United States

[2]Department of Computer Science, Dartmouth College, Hanover, NH, United States

[3]Department of Biomedical Data Science, Geisel School of Medicine, Dartmouth College, Lebanon, NH, United States

[4]Department of Psychiatry, Geisel School of Medicine, Dartmouth College, Lebanon, NH, United States

*Correspondence concerning this article should be addressed to Aiwei Zhang, Center for Technology and Behavioral Health, Dartmouth College, 46 Centerra Parkway Suite 300, Lebanon, NH 03766. Email: aiwei.zhang.25@dartmouth.edu

# Abstract

As wearable sensing becomes increasingly pervasive, a key challenge remains: how can we generate natural language summaries from raw physiological signals such as actigraphy - minute-level movement data collected via accelerometers? In this work, we introduce MotionTeller, a generative framework that natively integrates minute-level wearable activity data with large language models (LLMs). MotionTeller combines a pretrained actigraphy encoder with a lightweight projection module that maps behavioral embeddings into the token space of a frozen decoder-only LLM, enabling free-text, autoregressive generation of daily behavioral summaries.

We construct a novel dataset of 54,383 ⟨actigraphy, text⟩ pairs derived from real-world NHANES recordings, and train the model using cross-entropy loss with supervision only on the language tokens. MotionTeller achieves high semantic fidelity (BERTScore-F1 = 0.924) and lexical accuracy (ROUGE-1 = 0.722), outperforming prompt-based baselines by 7% in ROUGE-1. The average training loss converges to 0.38 by epoch 15, indicating stable optimization. Qualitative analysis confirms that MotionTeller captures circadian structure and behavioral transitions, while PCA plots reveal enhanced cluster alignment in embedding space post-training. Together, these results position MotionTeller as a scalable, interpretable system for transforming wearable sensor data into fluent, human-centered descriptions, introducing new pathways for behavioral monitoring, clinical review, and personalized health interventions.

# 1. Introduction

In recent years, the proliferation of wearable sensor devices has transformed how we monitor and understand human behavior in everyday settings. Actigraphy, which is minute-level activity traces captured via wrist-worn accelerometers, have become a core modality in both clinical and consumer health contexts. It provides a non-invasive, high-resolution view into physical movement and is widely used to assess sleep quality, circadian rhythm, depression, and medication adherence (Yoo et al. 2023; Dunn et al. 2021).

At the same time, large language models (LLMs) have revolutionized natural language understanding and generation, demonstrating strong generalization across diverse tasks through pretrained contextual representations (Naveed et al. 2024). Their applications in healthcare are expanding rapidly: LLMs have shown promise in generating medical explanations, supporting peer-based interventions, and simulating empathic conversation in mental health contexts (De Choudhury, Pendse, and Kumar 2023; Sharma et al. 2023). Yet as recent evaluations suggest, current LLMs often underperform on competencies most relevant to behavioral health (such as contextual sensitivity, emotional nuance, and ethical decision-making), particularly when applied to high-stakes counseling scenarios (Nguyen et al. 2025).

A key missing link is the ability to natively connect LLMs with behavioral sensor inputs such as actigraphy. Despite their complementary strengths, LLMs and actigraphy have traditionally existed almost completely independently. Sensor-based models, such as CNNs, RNNs, and transformers, have been used to classify behavioral states and predict health outcomes from raw actigraphy (Heinz et al. 2022; F. Ruan et al. 2024; Dorris, Oh, and Jacobson 2024), but they produce structured outputs like labels or scores instead of language. Moreover, many of these CNN and LSTM models rely on short time windows or handcrafted features and struggle to capture long-range dependencies in continuous, high-dimensional sequences (Rahman and Adjeroh 2019; Patterson et al. 2023).

Conversely, while LLMs are inherently generative, they lack the capacity to interpret raw time-series inputs directly. Attempts to bridge the gap between sensing and generation have typically relied on manual preprocessing or prompt engineering, such as tokenizing numeric sequences into digit-wise text strings or formatting prompts as question-answering tasks (Gruver et al. 2023; Kim et al. 2024; Nepal et al. 2024). These approaches often fragment the temporal structure of the original signal, leading to representations that are decoupled from the dynamics of real-world time-series data. Similarly, earlier interpretable models for clinical time-series, like RETAIN, require heavily structured and aggregated inputs, limiting flexibility when dealing with dense or high-frequency behavioral trajectories (Choi et al. 2016). This reliance on preprocessing and structure constrains the interpretability and adaptability of outputs, particularly in complex real-world contexts.

To address this, we introduce MotionTeller, a unified framework that aligns actigraphy and language through a generative pipeline. MotionTeller combines a pretrained transformer-based actigraphy encoder (PAT) (F. Y. Ruan et al. 2024), originally trained on over 29,000 participants from the NHANES dataset, with a lightweight projection module that maps sensor-derived embeddings into the token space of a frozen decoder-only LLM. This setup enables the model to condition autoregressive generation on raw activity traces and produce fluent, context-aware behavioral summaries. We also release a novel dataset of over 54,000 of ⟨raw sequence, generated label⟩ pairs, each consisting of a 24-hour actigraphy sequence and a GPT-generated summary crafted via few-shot prompting. Finally, we examine how semantic representations evolve across training and evaluate performance both quantitatively and qualitatively. Together, these contributions form a foundation for behavior-aware LLMs that translate physiological signals into human-readable narratives, offering new possibilities for personalized feedback, clinical interpretability, and scalable behavioral health tools.

# 2. Related Work

As large language models (LLMs) and wearable sensing technologies continue to evolve, a growing body of work explores how to connect time-series representations with language-based reasoning. Prior research spans traditional actigraphy modeling, health-oriented LLMs, early attempts at sensor-to-text generation, and architectural insights from multimodal foundation models. Yet, there remains a significant gap in directly aligning raw sensor data with autoregressive LLMs for free-form generation, which MotionTeller is designed to fill.

## 2.1 Actigraphy for Behavioral Modeling

Actigraphy has long been a core modality in health monitoring, particularly in studies of circadian rhythms, psychiatric disorders, and medication use. Traditional work relied on engineered features or statistical summaries over time windows to predict sleep patterns, depressive symptoms, or treatment adherence (Heinz et al. 2022). More recently, transformer-based models have shown promise in detecting behavioral states from minute-level activity traces (F. Y. Ruan et al. 2024). However, these models are fundamentally structured for classification or regression, producing scalar outputs rather than language. They offer limited interpretability beyond a numeric label and cannot explain why an activity pattern corresponds to a certain clinical state. Furthermore, most approaches operate on heavily preprocessed data, sacrificing the rich temporal granularity that actigraphy provides. To the best of our knowledge, no known method produces natural language descriptions directly from raw actigraphy, leaving a gap in making sensor data interpretable to non-technical users or downstream LLMs.

## 2.2. LLMs in Health and Behavioral Tasks

LLMs have demonstrated remarkable potential in healthcare applications, including clinical decision support, patient-centered dialogue, and mental health screening. Med-PaLM (Singhal et al. 2023) has shown that foundation models can achieve expert-level performance in medical question answering, while Me-LLaMA (Xie et al., 2024) illustrates the benefits of domain-specific pretraining for biomedical reasoning. In the mental health domain, LLMs are being explored not only for diagnosis and triage, but also for conversation-driven support. Recent work has validated the promise of generative technology in this space. Therabot, a fine-tuned LLM-based chatbot, demonstrated significant reductions in symptoms of depression, anxiety, and eating disorders in a randomized controlled trial (Heinz et al. 2025), with outcomes on par with human-delivered therapy across multiple engagement and alliance metrics. This provides early evidence that LLMs can support structured therapeutic interaction when fine-tuned on curated expert data. Related evaluations also highlight LLMs' strengths in capturing diagnostic structure while revealing variability in reasoning and performance across conditions (Heinz et al. 2023). However, these systems operate almost exclusively on linguistic inputs, including patient messages, transcripts, or clinical vignettes, and do not center on non-verbal behavioral signals such as actigraphy. MotionTeller builds on this emerging field by aligning LLMs with continuous, physiological input, treating sensor data not as metadata but as a primary source for free-form narrative generation.

## 2.3 Generating Text and Behavioral Narratives from Sensor Data

While LLMs have seen broad adoption in clinical natural language processing, the generation of human-readable summaries from raw sensor data remains underexplored. Prior work has either focused on classification (e.g., predicting conditions from wearables) or on purely prompting methods for basic description. Time2Lang (Pillai et al. 2025) made a significant step forward by directly aligning time-series embeddings with LLM token representations using a projection layer, hence reducing the dependency on handcrafted prompts and summary templates. HealthLLM (Kim et al. 2024) similarly proposes aligning wearable data with LLMs for prediction tasks but relies heavily on structured sensor statistics, rather than raw or high-resolution activity traces. By contrast, MotionTeller introduces a generative interface between a pretrained actigraphy encoder and a decoder-only LLM, allowing the model to interpret minute-level actigraphy data in its native form and produce rich, qualitative descriptions. This approach enables context-aware behavioral language generation that goes beyond simple pattern recognition or metadata tagging. On the other hand, JoLT (Cai et al. 2024) tackles a related problem by aligning ECG waveform embeddings with language using a Q-Former and a frozen OPT decoder. Their method supports both summarization and question answering but is designed for structured clinical time-series (e.g., ECG), and depends on text-aligned supervision during training. In contrast, MotionTeller uses behavioral signals like actigraphy, which lack structured annotations, and does not require paired expert-written summaries for training.

## 2.4 Prompting Limitations for Behavioral Sensing

Prompt-based generation has become a popular solution in domains with limited labeled data. In behavioral health, few-shot prompting with GPT models can produce fluent summaries, particularly when given structured inputs like sleep logs or step counts. However, as highlighted by prior evaluations (e.g., HealthLLM; Kim et al., 2024), these summaries often lack grounding in the data itself and are prone to producing generic or repetitive phrasing. MotionTeller addresses this by replacing prompt-based summarization with a learned alignment: a trainable projection module that embeds behavioral signals into the LLM's input space. As demonstrated in Section 5.1.4, this yields superior performance in both lexical and semantic metrics, especially for data types, like raw actigraphy, that lack intrinsic semantic anchors.

## 2.5 Alignment Strategies Across Modalities

The architecture of MotionTeller draws inspiration from recent advances in vision-language modeling. BLIP (J. Li et al. 2023; 2022) and Flamingo (Alayrac et al. 2022) demonstrate how frozen modality-specific encoders (e.g., vision) can be paired with frozen language decoders to enable multimodal generation through a trainable bridging layer. These models achieve strong results in captioning, visual QA, and storytelling by conditioning language models on non-text embeddings. MotionTeller adapts this architectural paradigm to wearable data: instead of images, we use high-frequency movement patterns; instead of visual attention maps, we rely on temporal patches learned via the Pretrained Actigraphy Transformer (PAT) (F. Y. Ruan et al. 2024). The projection module *f* in MotionTeller performs the equivalent role of vision-to-text alignment, except in a fundamentally different modality where spatial and semantic regularities are weaker. This approach allows MotionTeller to preserve the architectural modularity of foundation models while extending generative capacity to continuous behavioral signals. Another approach, ViTST (Z. Li, Li, and Yan 2023), tackles irregular time-series classification by converting signals into line-graph images and feeding them into pretrained vision transformers. While effective for classification and robust to missing data, this approach does not model temporal semantics or produce natural language outputs. MotionTeller differs by directly embedding raw time-series into language space for generative tasks, facilitating behavioral interpretation rather than prediction.

# 3. Methodology

## 3.1 The MotionTeller Dataset & Dataset Construction

### 3.1.1 NHANES Raw Actigraphy Data Overview

This study uses data from the 2013 to 2014 cohort of the National Health and Nutrition Examination Survey (NHANES) ("NHANES Homepage" 2024), a large, nationally representative health dataset managed by the Centers for Disease Control and Prevention. Each participant in this cohort was equipped with a triaxial wrist-worn accelerometer (ActiGraph GT3X+) that captured minute-level physical activity intensity over a continuous 7-day period, resulting in 10,080 data points per participant. These values represent the summed vector magnitude of movement at each minute. After initial participant filtering, the final dataset included 7,769 participants, each with a full week of valid actigraphy data. This yields a total of 54,383 individual 24-hour activity sequences, where each sequence consists of 1,440 raw minute-level values.

However, due to hardware constraints, all experiments in this study were conducted using a Google Colab Pro environment with a single NVIDIA L4 GPU using only one-seventh of the entire dataset, as it was not feasible to load and train on the full 7-day sequence per participant. Additionally, long input sequences impose considerable memory demands during sequence-to-sequence training, particularly when using autoregressive language models. As a result, this project focuses on a subset of the dataset, selecting one 24-hour sequence per participant, resulting in 7,769 ⟨raw sequence, generated label⟩ pairs used for training and evaluation in this thesis. In future work, we plan to scale up MotionTeller to leverage the full 54,383 ⟨raw sequence, generated label⟩ pairs available in the dataset, enabling multi-day modeling and more robust generalization across behavioral patterns.

### 3.1.2 Lack of Existing Paired Text Labels for Actigraphy

Neither the NHANES dataset nor any other publicly available actigraphy dataset includes paired natural language descriptions of daily activity. Existing actigraphy research typically focuses on classification tasks using structured categorical labels, such as medication use or sleep disorders, rather than descriptive, narrative summaries ("NHANES Homepage" 2024). To the best of our knowledge, there is no existing dataset that aligns raw time-series movement data with corresponding text. This limitation has hindered the development of generative models capable of producing human-readable behavioral summaries. This project addresses that gap by creating the first dataset of its kind: minute-level actigraphy sequences paired with GPT-generated daily activity summaries.

### 3.1.3 From Raw Signals to Structured Representations and Text Labels

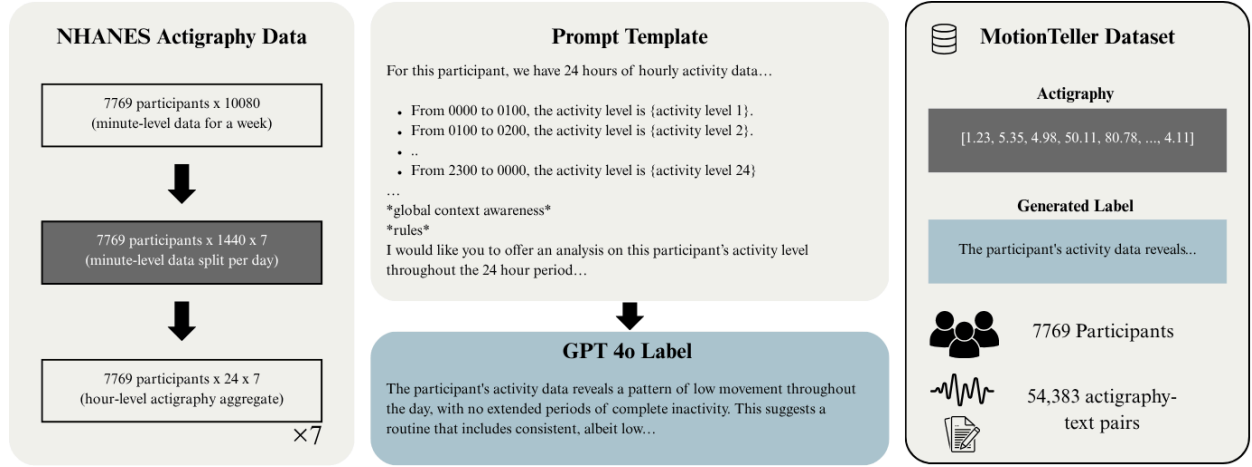

*Figure 1. **Pipeline for constructing the MotionTeller dataset.** Raw minute-level actigraphy data is first segmented into daily sequences and binned into 24 hourly activity levels. These structured inputs are used to prompt GPT-4o for generating human-readable summaries. The final dataset consists of ⟨raw sequence, generated label⟩ pairs, which are used to train MotionTeller to generate behavioral summaries directly from raw sensor data.*

To create the dataset used in this study, we developed a three-stage pipeline that bridges high-resolution actigraphy signals with natural language summaries. This process is illustrated in Figure 1 and consists of: (1) transforming raw minute-level signals into structured hourly profiles, (2) prompting a language model to generate text summaries from these profiles, and (3) pairing the generated summaries with the original raw sequences to form the final training dataset for MotionTeller.

**From Raw to Structured.** The source data consists of minute-level actigraphy traces from the 2013 to 2014 NHANES dataset, with each participant contributing 7 days of 10,080-minute sequences (i.e., 1,440 minutes/day × 7 days). Each day's data is split into a separate 24-hour segment, resulting in a total of 54,383 daily sequences across 7,769 participants. To reduce dimensionality while preserving circadian behavioral patterns, we bin each 24-hour sequence into 24 hourly values. This is done by computing the average movement intensity within each hour, yielding a compact and structured profile of daily activity. Each of these 24-dimensional vectors is then normalized globally across the dataset using min-max scaling, and further rescaled to integers between 0 and 1000. This preserves comparability across participants while making the input suitable for textual prompting.

**From Structure to Text.** The structured 24-hour profiles serve as input to a few-shot prompted GPT-4o model, which is instructed to act as a behavioral analyst. The prompt presents hourly activity levels using natural time references (e.g., "From 0000 to 0100, the activity level is {x}..."), followed by a set of instructions requesting a behavioral analysis of the full 24-hour period.

To construct high-quality few-shot examples, we first applied KMeans clustering on the 24-bin structured actigraphy data to identify five representative activity profiles that captured the diversity of movement patterns in the dataset. Each selected cluster centroid was matched with a real participant, whose daily activity sequence was then manually labeled to ensure semantic accuracy and interpretive clarity.

These hand-crafted ⟨structured activity, human-written summary⟩ pairs were then used as few-shot exemplars in the GPT-4o prompt, guiding the model to produce summaries that are consistent, fluent, and behaviorally grounded. The full prompting format, along with an example input and output, is provided in **Appendix A.1**.

**Constructing the MotionTeller Dataset.** After label generation, each participant's original 1,440-minute raw actigraphy sequence is paired with its corresponding GPT-generated summary, forming a ⟨raw sequence, generated label⟩ pair. These raw–text pairs constitute the input to MotionTeller. During training, the model learns to generate the textual summary directly from the raw minute-level sequence, leveraging the resolution of full actigraphy traces while benefiting from the linguistic grounding of the generated summaries. It is important to note that for evaluation purposes, while the summaries used for training are GPT-generated, the raw actigraphy sequence represents the authentic behavioral signal. As such, model evaluation focuses on whether generated text aligns with the input activity data, not with a potentially imperfect GPT label.

### 3.1.4 Label Verification and Dataset Finalization

To assess the quality of the GPT-generated activity summaries, we conducted a structured evaluation on a held-out set of 30 participants. These participants were selected from the NHANES 2003 to 2004 cohort, a distinct yet methodologically identical dataset with minute-level actigraphy data collected under the same protocol. This ensured that evaluation was not biased by data leakage from the training distribution.

To ensure diversity in activity patterns, we applied KMeans clustering to the structured 24-hour profiles and selected 6 participants from each of 5 clusters, producing a balanced evaluation set of 30 samples. This sample size is commonly used in cognitive science and machine learning literature as a statistically meaningful minimum for pilot validation of semantic fidelity and inter-rater reliability, while remaining feasible for detailed human review (Bujang et al. 2024).

The evaluation rubric was developed collaboratively and covered six distinct criteria:
1. Identification of Peak Activity
2. Description of Nighttime to Dawn (0000–0600)
3. Description of Early Morning to Noon (0600–1200)
4. Description of Afternoon Activity (1200–1800)
5. Description of Evening and End-of-Day (1800–0000)
6. Quality of Language and Bias Avoidance

Each criterion was scored on a 1 to 5 scale, where 5 represents perfect fidelity and 1 represents a significant error or omission. This yields a total possible score of 30 per participant summary. A full description of the rubric and score definitions is provided in Appendix B. The evaluation was conducted independently by two raters: the project first-author (Zhang) and a senior research mentor (Pillai). Neither rater communicated with the other during scoring. The average total scores were 28.3 (Zhang) and 28.6 (Pillai) out of 30, reflecting high quality and cross-rater agreement. Minor discrepancies occurred primarily in areas of stylistic preference or interpretive emphasis, not in core semantic accuracy.

Based on this evaluation, we concluded that the GPT-generated summaries were largely consistent, behaviorally faithful, and suitable for use as supervision targets in downstream model training.

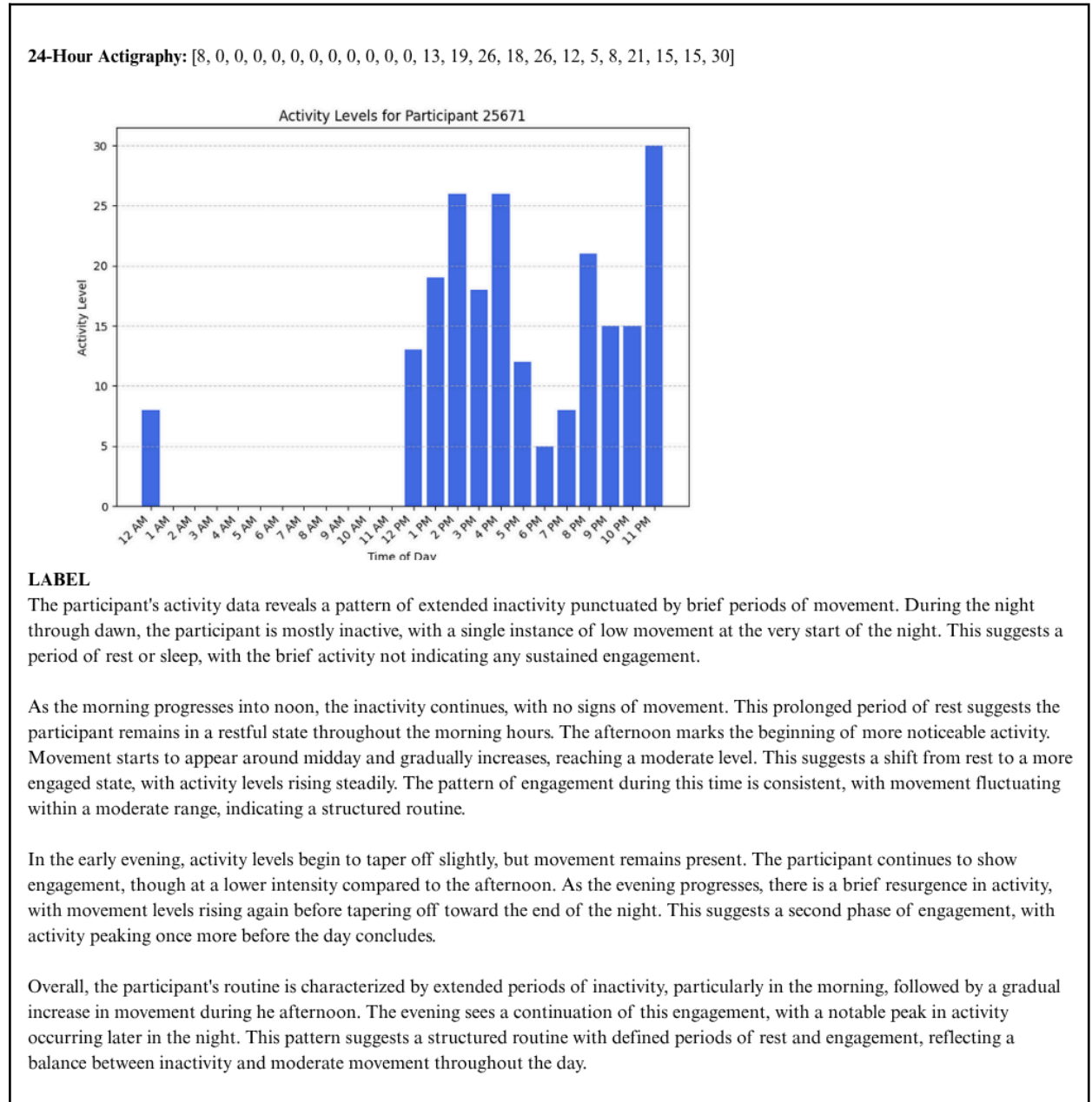

*Figure 2. Example of a 24-hour structured actigraphy sequence used as input to the GPT-4o label generation pipeline, shown alongside the resulting summary.* The bar chart displays hourly activity levels, and the summary reflects behavioral patterns across the night, morning, afternoon, and evening. This figure illustrates the kind of generated output evaluated using the rubric in Appendix B.

## 3.2 MotionTeller Model Architecture

MotionTeller consists of three major components: a frozen pretrained PAT encoder, a lightweight projection module, and a frozen decoder-only LLM.

### 3.2.1 Pretrained Actigraphy Transformer (PAT)

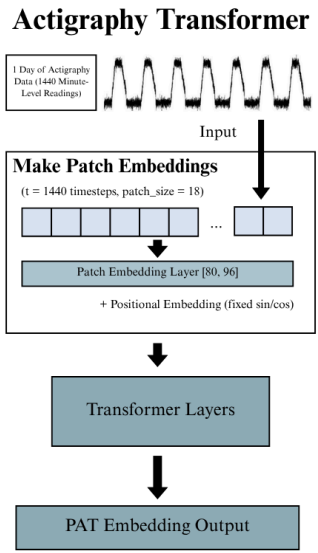

*Figure 3. Architecture of the pretrained Actigraphy Transformer (PAT).* *A single 1,440-minute actigraphy sequence is split into 80 patches of 18 minutes each. Each patch is embedded into a 96-dimensional vector and augmented with fixed sinusoidal positional encodings. The resulting sequence of patch tokens is passed through a stack of transformer encoder layers. The model outputs a contextualized embedding of shape [80, 96], which serves as the input to the MotionTeller projection module.*

The Pretrained Actigraphy Transformer (PAT) serves as the fixed encoder in the MotionTeller pipeline. Originally introduced in Ruan et al. (2024), PAT was pretrained on over 29,000 participants using a masked autoencoding objective and designed to encode daily actigraphy into semantically meaningful token representations.

For this study, we use the PAT-Large (PAT-L) variant, consisting of a patching layer, fixed sinusoidal positional encodings, and multiple transformer encoder blocks. Each raw actigraphy sequence of 1,440 minutes is split into 80 non-overlapping patches of 18 minutes. As shown in Figure 3, these patches are embedded and passed through the encoder, producing a final output of shape [80, 96]. The PAT model is loaded using its saved Keras .h5 weights and used strictly in inference mode, with all parameters frozen. For each participant-day, we pass the raw actigraphy vector into PAT and extract the resulting token-level embedding, which encodes both short- and long-range temporal dependencies in a compact form.

Finally, these [80, 96] PAT embeddings are then saved and passed to the downstream projection module (Section 3.2.2), which prepares them for input into the decoder-only language model. A detailed architectural schematic of PAT is presented in Figure 3. For full implementation specifics, such as encoder depth, pretraining strategy, and patching protocol, we refer readers to the original PAT publication by F. Y. Ruan et al. 2024. In the context of this work, PAT serves as a frozen, pretrained module that transforms high-resolution actigraphy into structured token embeddings suitable for alignment with a language model.

### 3.2.2 Projection Module *f*

The MotionTeller framework includes a lightweight projection module *f* that maps the pretrained PAT embeddings into the token embedding space expected by the decoder-only language model. Specifically, the output from PAT is a sequence of 80 embeddings of dimension 96, i.e., a shape of [80, 96] per participant-day. However, the decoder model (Gemma-2B) expects inputs of shape [N, 2048], where 2048 is the dimensionality of its token embedding space.

To bridge this mismatch, we define *f* as a simple feedforward neural network that transforms each of the 80 token vectors independently. It performs a linear projection from 96 to 2048 dimensions, effectively mapping [80, 96] → [80, 2048]. This projected output is then used as a contextual prefix to condition the decoder's autoregressive generation of behavioral summaries. While more complex architectures for *f*, such as 1D CNNs or LSTMs, were considered, empirical testing showed that they significantly increased inference time and training instability without improving downstream performance. In contrast, a simple fully connected layer (or two-layer MLP) was not only faster but also led to faster convergence and stable gradients during training. This simplicity enables efficient batch processing and ensures compatibility with frozen LLM architectures that are sensitive to input embedding shape.

The projection head *f* is the only trainable component in the MotionTeller pipeline. Both the PAT encoder and the LLM decoder remain frozen during training. As a result, the training objective focuses exclusively on learning a semantic projection from sensor-derived embeddings to the LLM's input space.

### 3.2.3 Decoder LLM (Gemma-2B) and Autoregressive Generation

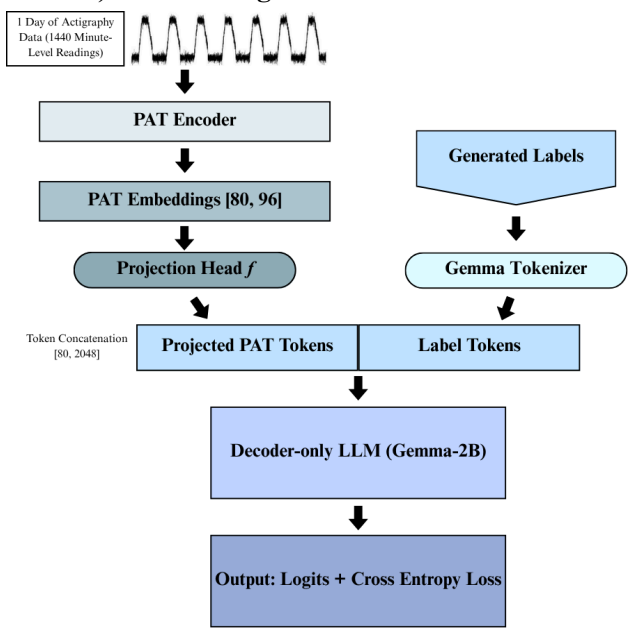

***Figure 4. Architecture of the MotionTeller generative pipeline.*** *Raw minute-level actigraphy is encoded into [80, 96] token embeddings using a frozen pretrained PAT encoder. These embeddings are projected into the decoder's embedding space ([80, 2048]) via a trainable projection head f. The projected actigraphy tokens are concatenated*

*with tokenized summary labels (via the Gemma tokenizer) and passed into a frozen decoder-only LLM (Gemma-2B). The model autoregressively generates the behavioral summary, and token-level cross-entropy loss is computed between predicted logits and the reference label tokens.*

The final component of the MotionTeller model is a decoder-only LLM that generates a behavioral summary conditioned on the projected actigraphy embeddings. For this component, we use the Gemma-2B model, a frozen decoder-only transformer trained for natural language generation. All weights in the LLM remain frozen during training.

As shown in Figure 4, the LLM receives a customized input: a concatenation of the projected PAT embeddings (shape [80, 2048]) and the token embeddings of the target summary, which are produced using the Gemma tokenizer. A special start-of-sequence token [BOS] is prepended to mark the beginning of the sequence. The full input sequence allows the decoder to condition on actigraphy data and generate the summary in an autoregressive manner.

During training, the attention mask is configured such that the model can attend to the entire PAT embedding prefix when predicting each token. The first 80 positions, which correspond to the projected PAT tokens, are excluded from loss computation by assigning them a label mask of -100. As a result, loss is only computed over the language portion of the sequence, while the PAT segment acts as a fixed prefix.

Although PAT embeddings and label tokens are structurally identical in shape (both 2048-dimensional vectors), they are semantically and positionally distinguishable. The model consistently receives the first 80 tokens as PAT embeddings, never associated with supervision. These tokens serve as a frozen, semantically rich context, akin to a prefix prompt. The decoder's causal attention mechanism ensures that each generated token is conditioned on both the PAT context and previously generated tokens. This setup is structurally analogous to prefix-based multimodal prompting, as seen in models like BLIP and Flamingo.

Finally, the decoder outputs a sequence of token-level logits, which are compared against the reference label tokens using cross-entropy loss. Only the projection head $f$ is updated during training; both the PAT encoder and the decoder LLM remain frozen. This modular architecture ensures parameter efficiency, semantic interpretability, and compatibility with large-scale pretrained models. All together, these components form a modular and interpretable pipeline for generating language from wearable time-series input.

At inference time, the model takes as input a raw actigraphy sequence from an unseen participant and generates a behavioral summary. The downstream task is thus defined as free-form generation conditioned on raw sensor data, with no label input at test time.

# 4. Experiments

The primary objective of MotionTeller is to generate fluent and behaviorally accurate textual summaries from raw wearable sensor data. During inference, the model receives only a raw 24-hour minute-level actigraphy sequence, which is passed through a frozen PAT encoder and projection head *f*, and uses the resulting token embeddings to condition the generation of a natural language summary. No label or reference text is provided at test time. The model must therefore learn to generalize from sensor-derived embeddings to coherent and semantically faithful language outputs. This section describes the experimental setup, training conditions, and evaluation procedures used to assess MotionTeller's performance on this downstream generative task.

## 4.1 Training Setup

All experiments were conducted using a Google Colab Pro + environment with a single NVIDIA L4 GPU, approximately 24 GB of RAM, and mixed-precision training enabled via PyTorch. The final dataset used in the actual experiment consisted of 7,769 ⟨raw sequence, generated label⟩ pairs, split into 80% training, 10% validation, and 10% test sets.

The model was trained for 15 epochs, each comprising over 3,000 training steps, resulting in a total runtime of more than 45 hours. A batch size of 2 was used for both training and validation due to GPU memory limitations. Training was conducted using teacher forcing and cross-entropy loss, applied only to the summary tokens (with the actigraphy prefix masked using a label value of -100; see Section 3.2.3).

A linear learning rate scheduler (LinearLR) was used to gradually warm up the learning rate over the first 100 steps. Training progress was logged every 10 steps, and model checkpoints were selected based on the validation BERTScore-F1. On average, each epoch required approximately 3 hours and 15 minutes, depending on runtime variability and logging overhead.

## 4.2 Evaluation Metrics

To assess the quality of MotionTeller's generated summaries, we use a combination of lexical overlap and semantic similarity metrics, applied to both the validation and test sets. These metrics evaluate whether the model can generate behavioral descriptions that are both linguistically accurate and semantically faithful to the GPT-generated reference summaries.

The following metrics are used for evaluation:
1. **ROUGE-1** and **ROUGE-L** measure word-level overlap between the generated summary (*prediction*) and the ground truth label (*reference*). ROUGE-1 captures unigram (individual word) overlap, while ROUGE-L measures the length of the longest common subsequence (LCS), which is especially useful for evaluating narrative structure and phrasing. These metrics were computed using the rouge_score library with stemming enabled, allowing for more flexible matching across different word forms (e.g., "sleep" vs. "sleeping").
2. **BERTScore (Precision, Recall, F1)** evaluates semantic similarity between the generated and reference summaries using contextual embeddings from a pretrained transformer. This allows the evaluation to account for paraphrasing and meaning preservation even when exact words do not match.

All metrics are computed on a per-participant basis and then macro-averaged across the dataset split. During testing, the model is evaluated on its ability to generalize on unseen actigraphy sequences from participants not present in the training set. These scores form the basis for the quantitative analyses presented in Section 5.1.

## 4.3 Cluster-Based Evaluation Subset

To ensure a structured and behaviorally diverse evaluation, we created a cluster-based subset of 100 participants drawn from the full dataset. We first applied KMeans clustering ($k = 5$) to the structured 24-hour actigraphy representations described in Section 3.1.3, which are used to generate labels in the MotionTeller Dataset. Each input consisted of 24 hourly activity values per participant, normalized to the [0, 1000] range. The clustering was performed on these 24-dimensional vectors to group participants based on coarse-grained behavioral archetypes, such as morning-dominant activity, low overall movement, or irregular temporal patterns.

From each of the five resulting clusters, we randomly sampled 20 participants, resulting in a balanced evaluation subset of 100 participants that reflects a broad spectrum of daily movement behaviors. This subset was used for multiple downstream purposes:

(a) In **Section 5.1.4** we use these participants to generate summaries via few-shot GPT-4o prompting directly on raw actigraphy, creating a baseline set of labels to compare against MotionTeller.
(b) In **Section 5.2**, we use an individual participant's result from this cohort in qualitative analysis, examining how well MotionTeller-generated summaries reflect ground-truth behavioral patterns.
(c) In **Section 5.3.1**, we report cluster-wise evaluation metrics using ROUGE and BERTScore, offering insight into the model's performance across distinct behavior types.

# 5. Results

This section presents both quantitative and qualitative evaluations of MotionTeller's performance. We begin with aggregate metrics on validation and test sets across multiple training epochs, followed by sample-level analysis of generated summaries. Finally, we evaluate MotionTeller against a prompting-based GPT-4o baseline to contextualize its generative capacity under minimal supervision.

## 5.1 Quantitative Performance

### 5.1.1 Validation Set Performance Across Training Epochs

We evaluated MotionTeller on the held-out validation set of 777 participants using ROUGE and BERTScore metrics at three checkpoints: after epochs 5, 10, and 15. These metrics assess the degree to which the model's generated summaries align with the GPT-generated reference labels, both lexically and semantically. Table X presents the mean and standard deviation of each metric across all participants.

*Table 1: Validation performance of MotionTeller on the full held-out test set (777 participants) at epochs 5, 10, and 15. Reported values are the mean and standard deviation across all participants. ROUGE metrics assess lexical and structural similarity to reference summaries, while BERTScore evaluates semantic alignment using contextual embeddings.*

| Epoch | 5 | 10 | 15 |
|---|---|---|---|
| **Metric** | **Mean ± Std Dev** | **Mean ± Std Dev** | **Mean ± Std Dev** |
| **Rouge-1** | 0.6837 ± 0.1407 | 0.7078 ± 0.1044 | 0.7224 ± 0.0634 |
| **Rouge-L** | 0.4147 ± 0.0970 | 0.4272 ± 0.0813 | 0.4294 ± 0.0664 |
| **BERTScore_P** | 0.9189 ± 0.0376 | 0.9216 ± 0.0241 | 0.9241 ± 0.0194 |
| **BERTScore_R** | 0.9120 ± 0.0312 | 0.9188 ± 0.0208 | 0.9219 ± 0.0137 |
| **BERTScore_F1** | 0.9154 ± 0.0342 | 0.9202 ± 0.0221 | 0.9230 ± 0.0161 |

As shown in Table 1, we observe steady improvements across all evaluation metrics throughout training.
- ROUGE-1 increases from 0.6837 ± 0.1407 at epoch 5 to 0.7224 ± 0.0634 at epoch 15, indicating better lexical overlap with reference summaries.
- ROUGE-L improves similarly, suggesting that the model increasingly captures the structural and sequential elements of the narrative.
- BERTScore-F1 rises from 0.9154 ± 0.0342 to 0.9230 ± 0.0161, confirming the model's enhanced semantic alignment with the reference texts.

Remarkably, while the gains after epoch 10 are relatively modest, the projection module $f$ receives over 3,000 gradient updates per epoch, enabling it to converge quickly. Since the PAT encoder and decoder LLM are both frozen, this efficient adaptation through $f$ reflects its effectiveness in aligning actigraphy embeddings with the LLM's input space. The narrowing standard deviations across epochs also suggest increased consistency in summary quality across participants.

These validation results confirm that MotionTeller is not only able to generate summaries with high lexical and semantic fidelity, but that its performance stabilizes within the first half of training, supporting a parameter-efficient generative pipeline for wearable sensor data.

### 5.1.2 Training Loss Across Epochs

To better understand the model's convergence behavior, we tracked the average training loss per epoch across all 15 epochs. The results are visualized in Figure 5, which plots the loss curve during fine-tuning of the projection module $f$.

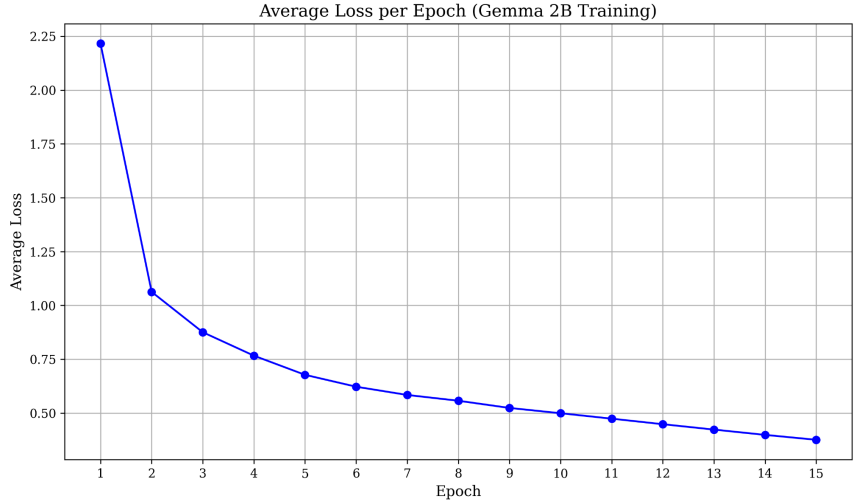

*Figure 5: Average training loss per epoch during fine-tuning of the projection module f over 15 epochs. The model converges rapidly within the first few epochs and continues to improve gradually thereafter.*

As shown in Figure 5, MotionTeller's training loss exhibits a rapid and stable decline, dropping from 2.22 in epoch 1 to 0.38 by epoch 15. The most significant decrease occurs within the first five epochs, after which the loss continues to decline gradually. This trend aligns with the observed plateau in validation performance (Section 5.1.1), and supports the interpretation that the model converges quickly due to the high frequency of updates, over 3,000 training steps per epoch, despite $f$ being the only trainable module.

There is no visible trend of overfitting or instability. The loss curve indicates stable optimization, reinforcing the viability of MotionTeller's architecture for efficient training in low-resource environments.

### 5.1.3 Test Set Performance

To evaluate MotionTeller's ability to generalize to unseen participants, we assessed its performance on a held-out test set of 777 participants, using the same metrics and checkpoints (epochs 5, 10, and 15) as in the validation analysis. The results are presented in Table 2.

*Table 2: Test set performance of MotionTeller on 777 unseen participants at epochs 5, 10, and 15. Reported values are the mean and standard deviation across all participants. The results show close alignment with validation performance (see Table 1), indicating strong generalization to unseen actigraphy inputs.*

| Epoch | 5 | 10 | 15 |
|---|---|---|---|
| **Metric** | **Mean ± Std Dev** | **Mean ± Std Dev** | **Mean ± Std Dev** |
| **Rouge-1** | 0.6843 ± 0.1428 | 0.7114 ± 0.0990 | 0.7221 ± 0.0614 |
| **Rouge-L** | 0.4137 ± 0.0974 | 0.4286 ± 0.0798 | 0.4330 ± 0.0655 |
| **BERTScore_P** | 0.9185 ± 0.0384 | 0.9222 ± 0.0230 | 0.9249 ± 0.0158 |
| **BERTScore_R** | 0.9119 ± 0.0318 | 0.9192 ± 0.0195 | 0.9222 ± 0.0129 |
| **BERTScore_F1** | 0.9152 ± 0.0349 | 0.9207 ± 0.0209 | 0.9235 ± 0.0137 |

Across all three epochs, test set performance closely aligns with validation performance. For example, ROUGE-1 improves from 0.6843 ± 0.1428 (epoch 5) to 0.7221 ± 0.0614 (epoch 15), nearly identical to the validation trajectory as shown in Table 1. Similar trends are observed in ROUGE-L and all BERTScore metrics, with BERTScore-F1 increasing from 0.9152 ± 0.0349 to 0.9235 ± 0.0137. This high degree of consistency suggests that MotionTeller is not only stable during training but also capable of retaining performance across distributional shifts, even when generating summaries for entirely unseen actigraphy sequences.

These results provide further evidence that the projection module $f$ learns a robust and generalizable alignment between sensor embeddings and language space. Because no label information is available at test time, the model's strong performance confirms that it is not overfitting to training set patterns or label style. Instead, it has learned to condition on raw actigraphy alone and generate fluent, semantically aligned summaries across a wide range of behavioral profiles.

The small and consistently low standard deviations across metrics also point to generally uniform performance across participants, indicating that MotionTeller is not just strong on average, but also dependable across individuals with diverse activity patterns.

### 5.1.4 Prompting-Base Baseline Comparison

To evaluate whether MotionTeller meaningfully improves over direct prompting on raw actigraphy, we compare its outputs to a baseline condition in which GPT-4o is prompted using few-shot examples derived from raw minute-level sequences, as first described in Section 4.3. This baseline uses the same five hand-labeled ⟨raw sequence, generated label⟩ pairs for few-shot prompting as described in Section 4.3, but relies entirely on zero pretraining or learned representations. The resulting summaries were then scored against the MotionTeller Dataset labels (generated using structured hourly data) using the same evaluation metrics.

*Table 3: Comparison of prompting-based GPT-4o baseline (raw input + few-shot) with MotionTeller's output after 15 epochs on the same 100-person evaluation subset. MotionTeller outperforms the baseline across all metrics, confirming the benefit of learning a structured embedding space over relying on prompt engineering alone.*

| Metric | Raw Baseline | MotionTeller Output (15 epochs) |
|---|---|---|
| **Rouge-1** | 0.6453 ± 0.0800 | 0.7221 ± 0.0614 |
| **Rouge-L** | 0.3425 ± 0.0522 | 0.4330 ± 0.0655 |
| **BERTScore_P** | 0.9028 ± 0.0150 | 0.9249 ± 0.0158 |
| **BERTScore_R** | 0.9084 ± 0.0135 | 0.9222 ± 0.0129 |
| **BERTScore_F1** | 0.9056 ± 0.0139 | 0.9235 ± 0.0137 |

Table 3 presents the comparison between the raw baseline and the MotionTeller output at epoch 15 on the same 100-person cluster-based evaluation subset. MotionTeller outperforms the prompting baseline across all metrics, with particularly large gains in ROUGE-1 (~7.7%) and ROUGE-L (~9%), indicating substantial improvement in both lexical overlap and structural alignment. Semantic metrics also improved, with BERTScore-F1 rising from 0.9056 to 0.9235, reflecting higher-quality language that is more faithful to the intended behavioral summaries.

These baseline results demonstrate that MotionTeller's learned alignment between actigraphy and language space, enabled by the PAT encoder and projection head, offers significant performance advantages over non-trainable prompt-based approaches. While prompting alone can produce broadly readable summaries, it fails to capture participant-specific behavioral nuance and often reverts to generic phrasing. MotionTeller, in contrast, leverages full-resolution input and token-level conditioning to produce semantically richer and more behaviorally grounded outputs.

### 5.1.5 Quantitative Performance Summary

In summary, the results across validation and test sets demonstrate that MotionTeller is capable of generating high-quality behavioral summaries from raw actigraphy sequences with both semantic fidelity and lexical structure. The projection module $f$ converges efficiently within the first few epochs, while performance remains stable and consistent through epoch 15. The strong alignment between validation and test metrics confirms that MotionTeller generalizes well to unseen participants, supporting its use as a lightweight, interpretable, and robust model for language-based behavioral understanding in wearable sensing applications.

## 5.2 Qualitative Performance

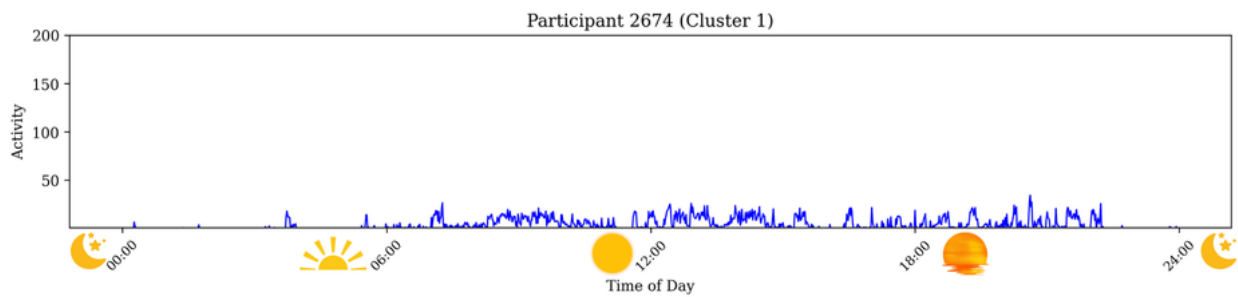

*Figure 6. Example of a MotionTeller-generated behavioral summary for a full-day actigraphy sequence. The bar chart (see above) visualizes minute-level physical activity data across a full day, where each bar represents movement intensity per minute. The accompanying summary (see below) is generated by MotionTeller and segmented into behavioral descriptions spanning midnight to dawn, dawn to noon, noon to evening, and evening to midnight. This example is drawn from a representative participant in the 100-person cluster-based evaluation subset and demonstrates how MotionTeller captures both temporal dynamics and overall behavioral patterns with natural language. Key statements in the summary reflect global reasoning (e.g., recognizing consistent low-level engagement), local event alignment (e.g., brief morning activity), and narrative structure.*

| Section | Generated Label |
|---|---|
| **Overall Statement** | "The participant's activity data reveals a pattern of low movement throughout the day, with activity levels remaining within a narrow range. This suggests that the participant may have a routine that involves a significant amount of low-level engagement, with some fluctuations in activity levels. |
| **Midnight (00:00) to Dawn (06:00) Description** | During the night to dawn, the participant shows minimal movement, indicating a period of rest or sleep. There are brief moments of activity, but these are sporadic and do not suggest a continuous period of engagement. |
| **Dawn (06:00) to Noon (12:00) Description** | As the morning progresses, activity levels increase slightly, indicating a transition from rest to a more active state. The movement remains low, but there is a gradual rise, suggesting a gradual increase in engagement as the day unfolds. |
| **Noon (12:00) to Late Afternoon (18:00) Description** | In the afternoon, the pattern of low movement continues, with activity levels remaining consistent and steady. There is no significant increase or decrease, indicating a sustained period of low-level engagement. This suggests a routine that involves regular, low-intensity activity. |
| **Evening (18:00) to Midnight (24:00) Description** | During the early evening, the participant maintains a similar pattern of low movement. Activity levels remain consistent and do not show any significant fluctuations. This suggests a continuation of the participant's routine, with no dramatic changes in engagement. |
| **Summarizing Statement** | Overall, the participant's activity data suggests a routine characterized by low movement, with a gradual increase in activity levels from morning to evening. This pattern may reflect a structured daily routine, with consistent but low-intensity engagement throughout the day." |

As shown in the example above, the participant's activity profile shows a clear low-intensity period during the early morning hours, followed by a gradual increase in movement across the day, peaking modestly in the evening. The generated summary accurately reflects these trends in multiple ways. We can examine a few details:

- First, the sentence "The participant's activity data reveals a pattern of low movement throughout the day, with activity levels remaining within a narrow range." (highlighted in blue) demonstrates MotionTeller's ability to reason about global context across the full day, rather than over-indexing on local peaks or noise.
- Second, the phrase "There are brief moments of activity..." (green) corresponds to small spikes in the morning seen between 0600 and 0900, which are faithfully echoed in the language output without exaggeration.
- Third, the clause "There is no significant increase or decrease, indicating a sustained period of low-level engagement" (grey) matches the pattern in the actigraphy trace from midday to evening, showing the model's ability to describe progressive behavioral arcs.
- These details illustrate not just the model's semantic correctness, but its ability to anchor description in behaviorally meaningful segments, striking a balance between high-level narrative framing and faithful signal interpretation.

However, it is important to acknowledge that some of the language remains relatively generic, a known tendency of decoder-only LLMs. For instance, phrases like "a gradual rise in engagement" or "consistent low-level activity" are behaviorally plausible but could also be applied, without much revision, to other participants with similar patterns. This highlights a limitation in MotionTeller's ability to produce participant-specific, discriminative summaries, despite strong alignment to the general temporal structure. Improving the precision and distinctiveness of generated text remains an important future direction.

Note that NHANES actigraphy values represent minute-level activity intensities based on MIMS (Monitor-Independent Movement Summary) units, which are device-independent and unitless. While not directly interpretable as physical quantities like steps or energy expenditure, they provide a reliable estimate of movement magnitude per minute. For more details on interpreting NHANES accelerometer data, please refer to **Appendix B: Understanding MIMS Units and Movement Intensity Thresholds**.

## 5.3 Representative Analysis

To analyze how MotionTeller reshapes the internal representation of actigraphy data, we performed principal component analysis (PCA) on the participant-level embeddings. Specifically, we compared the original embedding output from the frozen PAT encoder to the projected embeddings produced by MotionTeller (after the projection module $f$) at epoch 15. For each participant, we averaged the 80 token embeddings along the time dimension to obtain a single vector per participant (i.e., shape [96] for PAT, [2048] for MotionTeller). This allowed us to reduce the data to 2D for visualization using PCA.

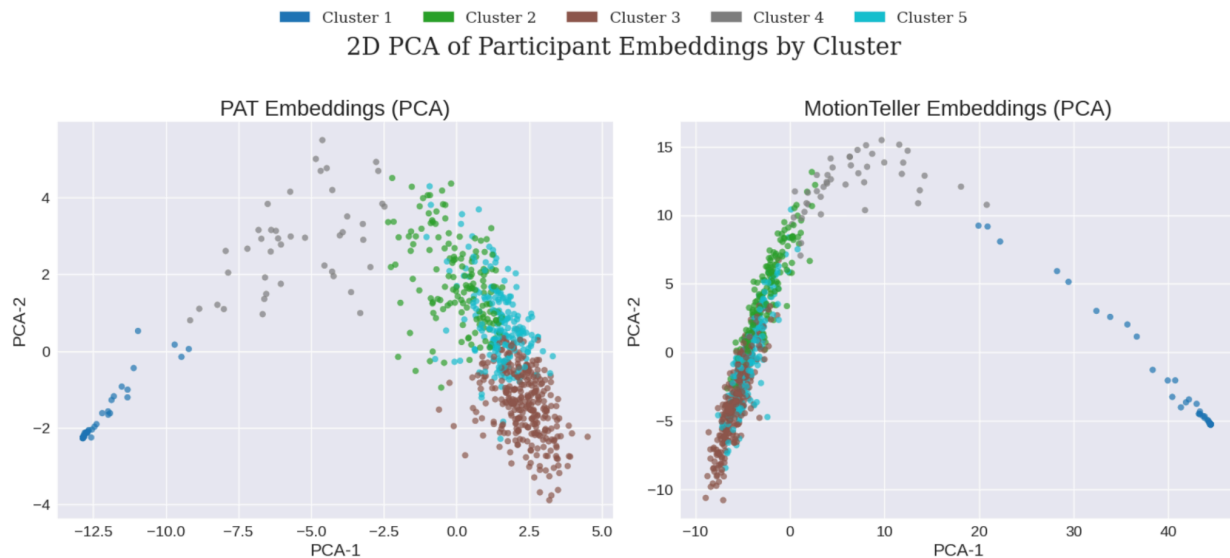

*Figure 7: 2D PCA of participant embeddings before and after MotionTeller training.* Left: original PAT embeddings (frozen). Right: MotionTeller embeddings after projection by f. Each dot represents a participant, color-coded by behavior cluster (dark blue = Cluster 1, green = Cluster 2, brown = Cluster 3, grey = Cluster 4, turquoise = Cluster 5).

PCA identifies the two orthogonal directions of greatest variance in the data. The x-axis (PCA-1) captures the most significant axis of variability across participants' embeddings, while the y-axis (PCA-2) captures the second-most significant. While the absolute axes are not directly interpretable in semantic terms, relative position and spatial separation reveal the internal geometry of the embedding space, showing how participants are organized by the model.

In the left panel of Figure 7, we visualize the frozen PAT embeddings. The clusters appear somewhat diffuse and overlapping, particularly between Clusters 2, 4, and 5. The right panel shows the embeddings after transformation by MotionTeller's projection module. Here, we observe sharper alignment of points along a shared axis and increased local cohesion within clusters, particularly Clusters 3 and 4. This suggests that the projection module *f* has learned to organize behaviorally similar participants into semantically meaningful regions of the latent space, better suited for downstream generation.

This structural reorganization provides evidence that MotionTeller's representation learning is not only functionally successful (as shown by evaluation metrics), but also geometrically interpretable, aligning embeddings according to shared behavioral traits.

### 5.3.1 Cluster-Wise Model Performance

To further investigate how MotionTeller performs across different behavioral types, we computed evaluation metrics for each of the five clusters described in Section 4.3, using the subset of 100 participants (20 per cluster). Results are shown in Table 4.

*Table 4. Cluster-wise evaluation of MotionTeller-generated summaries at epoch 15. Each column represents a distinct behavioral cluster (defined via KMeans on 24-hour activity profiles), and rows show average performance across 20 participants per cluster (100 total). Metrics include ROUGE-1 and ROUGE-L for lexical and structural overlap, and BERTScore F1 for semantic alignment with reference summaries. MotionTeller achieves the highest performance on Cluster 4, which likely represents participants with structured, interpretable behavioral patterns. In contrast, Cluster 1, characterized by lower and more ambiguous activity—shows consistently lower scores across all metrics*

| Metrics | Cluster 1 | Cluster 2 | Cluster 3 | Cluster 4 | Cluster 5 |
|---|---|---|---|---|---|
| **Rouge 1** | 0.6460 ± 0.0777 | 0.7216 ± 0.0497 | 0.7438 ± 0.0317 | **0.7499 ± 0.0163** | 0.7362 ± 0.0319 |
| **Rouge L** | 0.3697 ± 0.0512 | 0.4283 ± 0.0594 | 0.4432 ± 0.0612 | **0.4715 ± 0.0504** | 0.4478 ± 0.0560 |
| **BERTScore F1** | 0.9118 ± 0.0114 | 0.9227 ± 0.0129 | 0.9250 ± 0.0109 | **0.9312 ± 0.0072** | 0.9254 ± 0.0106 |

In contrast, Clusters 3 to 5 (brown, turquoise, and grey) exhibit both tighter embedding distributions and higher evaluation scores, with Cluster 4 achieving the highest BERTScore-F1 (0.9312), ROUGE-1 (0.7499) and ROUGE-L (0.4715). These clusters likely correspond to richer, more structured behavior profiles (e.g., activity spikes or circadian regularity), which are more amenable to consistent summarization.

These trends suggest that the semantic separability of participant embeddings after training, visible in PCA, correlates with the model's downstream performance across behavior types. Clusters with well-aligned embeddings tend to support higher-quality text generation, reinforcing the effectiveness of MotionTeller's projection-based alignment.

# 6. Discussion

This section reflects on the implications, strengths, and limitations of the MotionTeller architecture. We discuss the model's performance across diverse behavior types, its potential for scalable behavioral monitoring, and key areas for improvement including grounding, explainability, and clinical relevance.

## 6.1 Interpretations of the MotionTeller Framework

MotionTeller demonstrates that it is possible to align raw behavioral signals with natural language through a lightweight, modular architecture. Across multiple quantitative metrics, which include ROUGE, BERTScore, and cluster-specific evaluations, the model achieves considerably high semantic and lexical fidelity, even without finetuning the decoder LLM. Notably, the model performs best on clusters with structured activity patterns, suggesting that it learns to anchor behavioral variation within a coherent narrative space.

The success of the projection module $f$, despite its simplicity, highlights the strength of modular cross-modal design: raw actigraphy embeddings, when aligned with token representations, become expressive enough to support rich, autoregressive language generation. This is further validated by the PCA analysis of representation space before and after training, which shows a shift toward tighter, semantically organized clusters. Importantly, this organization emerges without supervised labels, indicating that MotionTeller learns to map actigraphy into meaningful latent regions of the LLM's input space using only generative supervision.

Qualitative analysis also reveals that the model reliably captures broad behavioral rhythms and transitions, with outputs that reflect daily structure and movement intensity. At the same time, some outputs rely on generalized phrasing, underscoring both the strength and the limitations of LLMs trained on relatively homogeneous summaries. All in all, MotionTeller bridges a novel representational gap, unlocking natural language generation from low-semantic, temporally continuous sensor input.

## 6.2 Implications for Future Health Applications

MotionTeller's generative approach to behavioral data unlocks a new class of applications in digital health and human-computer interaction. By producing context-aware, human-readable summaries from raw actigraphy, the model offers an interpretable interface between movement and meaning. These summaries could be embedded into patient-facing dashboards, used for therapeutic journaling, or support clinicians in reviewing behavioral trends over time.

Because MotionTeller is modular - one that uses a frozen encoder and decoder - it offers a flexible foundation for further adaptation. Different encoder architectures (e.g., for multimodal input) or decoder types (e.g., instruction-tuned models) could be integrated with minimal changes. This design supports scalability and personalization, while maintaining transparency and separation of concerns between signal encoding and linguistic reasoning.

More broadly, MotionTeller demonstrates that LLMs can serve not just as linguistic generators, but as semantic translators for physiological signals. As wearable technologies become more ubiquitous, systems like MotionTeller could help bridge the gap between raw behavioral data and human-centered explanation, eventually achieving scalable, personalized, and interpretable behavioral health tools.

## 6.3 Limitations and Areas for Improvement

While promising, MotionTeller has several limitations that offer direction for future work. First, all supervision comes from GPT-generated summaries, which, despite being fluent and behaviorally grounded, lack diversity in phrasing and stylistic nuance. As a result, the model may learn to mimic structural patterns without fully exploring alternative modes of description. Improving label expressiveness, perhaps through curated few-shot prompts or fine-tuned labeling models, could enhance generative richness.

Second, the frozen decoder simplifies training and reduces parameter count, but it may limit personalization or linguistic adaptability. In settings that require domain-specific tone or more varied conversational output, finetuning the decoder or adding adapter layers may offer value. Additionally, some activity traces are ambiguous or context-sensitive: a flat trace could indicate sleep, rest, or non-wear. Without additional context, the model must guess, subsequently raising challenges for interpretation and reliability.

Finally, the current system is unidirectional: it summarizes behavior but cannot yet support interactive querying or clarification. Extending the system to support question answering or feedback-driven refinement would make it more usable in real-world clinical or personal health settings. Further, incorporating complementary signals such as self-reported mood, context diaries, or sleep logs could help disambiguate activity traces and support more grounded, multimodal interpretation.

# 7. Conclusion

In this thesis, we proposed MotionTeller, a novel framework for generating behavioral summaries directly from raw wearable sensor data using a pretrained actigraphy encoder and a frozen large language model (LLM). By combining a frozen PAT encoder, a lightweight trainable projection module, and a frozen decoder-only LLM, MotionTeller is able to align high-resolution time-series data with natural language in an efficient and semantically faithful manner. We introduced a new actigraphy-text dataset, designed an effective training and evaluation pipeline, and demonstrated through both quantitative and qualitative evaluations that MotionTeller produces fluent, contextually relevant summaries grounded in behavioral structure.

Our results show that MotionTeller consistently outperforms prompting-based baselines, generalizes to unseen participants, and produces interpretable outputs that reflect both temporal and semantic alignment. The architecture is efficient, adaptable, and opens new possibilities for using LLMs in behavioral modeling tasks.

## Future Work
There are several promising directions for expanding this work:
1. **Scaling up data and granularity**: Future iterations could leverage multi-day actigraphy sequences to support longitudinal behavior tracking and include more granular labeling, such as splitting raw data into four-hour segments and attaching fine-grained text descriptions every 10 to 15 minutes. This could increase interpretability and enable richer narrative modeling.
2. **Improving the variety and richness of label supervision**: The current GPT-generated summaries, while fluent and structurally consistent, often exhibit lexical homogeneity and template-like phrasing (e.g., repeated use of patterns such as "the participant's activity is characterized by..."). This may limit the expressive range of the model's outputs. Future work could focus on enhancing the linguistic richness of training labels through better-curated few-shot examples, alternative prompting strategies, or the use of different LLMs beyond GPT-4o. This would promote greater stylistic variation and behavioral nuance in generated summaries and reduce the tendency toward generic or overgeneralized language.
3. **Fine-tuning the decoder LLM**: Currently, the LLM remains frozen. Fine-tuning it jointly with the projection head, potentially in a lightweight adapter-style manner, could improve fluency, specificity, and style alignment with behavioral language.
4. **Interactive downstream tasks**: Beyond free-form generation, MotionTeller could be extended to support interactive questions and answering on actigraphy sequences, such as answering queries like "Was there a period of restlessness during sleep?" or "At what point did the participant's activity peak?" This would allow richer clinical or user-facing applications.
5. **Multimodal integration**: Incorporating additional data sources (such as sleep diaries, mood logs, or environmental context) could help bridge gaps between observed behavior and subjective experience, strengthening both personalization and real-world relevance.

As wearable data becomes more ubiquitous and LLMs continue to evolve, MotionTeller offers a flexible, interpretable, and scalable foundation for translating behavioral signals into meaningful, human-centered language. Alongside this architectural contribution, this work introduces the MotionTeller Dataset - a novel, structured corpus of actigraphy-aligned behavioral summaries - which enables training and evaluation at scale. Through rigorous quantitative benchmarks and carefully curated qualitative analysis, MotionTeller demonstrates its ability to generalize across diverse participants, behaviors, and cluster types, establishing a new standard for text generation from time-series data.

More broadly, this work illustrates the untapped potential of generative foundation models in health and behavioral modeling. By bridging the gap between raw physiological signals and expressive language, MotionTeller paves the way for future systems that can explain, interact with, and even anticipate human behaviors across clinical, wellness, and daily-life contexts. It represents an important step toward the integration of LLMs into personalized, data-driven health technology, which is a frontier at the intersection of computational methods, behavioral science, and human-centered care.

# References

Alayrac, Jean-Baptiste, Jeff Donahue, Pauline Luc, Antoine Miech, Iain Barr, Yana Hasson, Karel Lenc, et al. 2022. "Flamingo: A Visual Language Model for Few-Shot Learning." In *Proceedings of the 36th International Conference on Neural Information Processing Systems*. NIPS '22. Red Hook, NY, USA: Curran Associates Inc.

Bujang, Mohamad Adam, Evi Diana Omar, Diana Hui Ping Foo, and Yoon Khee Hon. 2024. "Sample Size Determination for Conducting a Pilot Study to Assess Reliability of a Questionnaire." *Restorative Dentistry & Endodontics* 49 (1): e3. https://doi.org/10.5395/rde.2024.49.e3.

Cai, Yifu, Arvind Srinivasan, Mononito Goswami, Arjun Choudhry, and Artur Dubrawski. 2024. "JoLT: Jointly Learned Representations of Language and Time-Series for Clinical Time-Series Interpretation (Student Abstract)." *Proceedings of the AAAI Conference on Artificial Intelligence* 38 (21): 23447–48. https://doi.org/10.1609/aaai.v38i21.30423.

Choi, Edward, Mohammad Taha Bahadori, Joshua A. Kulas, Andy Schuetz, Walter F. Stewart, and Jimeng Sun. 2016. "RETAIN: An Interpretable Predictive Model for Healthcare Using Reverse Time Attention Mechanism." In *Proceedings of the 30th International Conference on Neural Information Processing Systems*, 3512–20. NIPS'16. Red Hook, NY, USA: Curran Associates Inc.

Chomistek, Andrea K., Changzheng Yuan, Charles E. Matthews, Richard P. Troiano, Heather R. Bowles, Jennifer Rood, Junaidah B. Barnett, Walter C. Willett, Eric B. Rimm, and David R. Bassett. 2017. "Physical Activity Assessment with the ActiGraph GT3X and Doubly Labeled Water." *Medicine & Science in Sports & Exercise* 49 (9): 1935–44. https://doi.org/10.1249/MSS.0000000000001299.

De Choudhury, Munmun, Sachin R. Pendse, and Neha Kumar. 2023. "Benefits and Harms of Large Language Models in Digital Mental Health." arXiv. https://doi.org/10.48550/ARXIV.2311.14693.

Dorris, Hannah, Jenny Oh, and Nicholas Jacobson. 2024. "Wearable Movement Data as a Potential Digital Biomarker for Chronic Pain: An Investigation Using Deep Learning." *Physical Activity and Health* 8 (1): 83–92. https://doi.org/10.5334/paah.329.

Dunn, Jessilyn, Lukasz Kidzinski, Ryan Runge, Daniel Witt, Jennifer L. Hicks, Sophia Miryam Schüssler-Fiorenza Rose, Xiao Li, et al. 2021. "Wearable Sensors Enable Personalized Predictions of Clinical Laboratory Measurements." *Nature Medicine* 27 (6): 1105–12. https://doi.org/10.1038/s41591-021-01339-0.

Evenson, Kelly R, and Fang Wen. 2015. "Performance of the ActiGraph Accelerometer Using a National Population-Based Sample of Youth and Adults." *BMC Research Notes* 8 (1): 7. https://doi.org/10.1186/s13104-014-0970-2.

Gruver, Nate, Marc Finzi, Shikai Qiu, and Andrew G Wilson. 2023. "Large Language Models Are Zero-Shot Time Series Forecasters." In *Advances in Neural Information Processing Systems*, edited by A. Oh, T. Naumann, A. Globerson, K. Saenko, M. Hardt, and S. Levine, 36:19622–35. Curran Associates, Inc. https://proceedings.neurips.cc/paper_files/paper/2023/file/3eb7ca52e8207697361b2c0fb3926511-Paper-Conference.pdf.

Heinz, Michael V., Sukanya Bhattacharya, Brianna Trudeau, Rachel Quist, Seo Ho Song, Camilla M. Lee, and Nicholas C. Jacobson. 2023. "Testing Domain Knowledge and Risk of Bias of a Large-Scale General Artificial Intelligence Model in Mental Health." *DIGITAL HEALTH* 9 (January):20552076231170499. https://doi.org/10.1177/20552076231170499.

Heinz, Michael V., Daniel M. Mackin, Brianna M. Trudeau, Sukanya Bhattacharya, Yinzhou Wang, Haley A. Banta, Abi D. Jewett, Abigail J. Salzhauer, Tess Z. Griffin, and Nicholas C. Jacobson. 2025. "Randomized Trial of a Generative AI Chatbot for Mental Health Treatment." *NEJM AI* 2 (4). https://doi.org/10.1056/AIoa2400802.

Heinz, Michael V., George D. Price, Franklin Ruan, Robert J. Klein, Matthew Nemesure, Aliza Lopez, and Nicholas C. Jacobson. 2022. "Association of Selective Serotonin Reuptake Inhibitor Use With Abnormal Physical Movement Patterns as Detected Using a Piezoelectric Accelerometer and Deep Learning in a Nationally Representative Sample of Noninstitutionalized Persons in the US." *JAMA Network Open* 5 (4): e225403. https://doi.org/10.1001/jamanetworkopen.2022.5403.

Kim, Yubin, Xuhai Xu, Daniel McDuff, Cynthia Breazeal, and Hae Won Park. 2024. "Health-LLM: Large Language Models for Health Prediction via Wearable Sensor Data." In *Proceedings of the Fifth Conference on Health, Inference, and Learning*, edited by Tom Pollard, Edward Choi, Pankhuri Singhal, Michael Hughes, Elena Sizikova, Bobak Mortazavi, Irene Chen, et al., 248:522–39. Proceedings of Machine Learning Research. PMLR. https://proceedings.mlr.press/v248/kim24b.html.

Li, Junnan, Dongxu Li, Silvio Savarese, and Steven Hoi. 2023. "BLIP-2: Bootstrapping Language-Image Pre-Training with Frozen Image Encoders and Large Language Models." In *Proceedings of the 40th International Conference on Machine Learning*, edited by Andreas Krause, Emma Brunskill, Kyunghyun Cho, Barbara Engelhardt, Sivan Sabato, and Jonathan Scarlett, 202:19730–42. Proceedings of Machine Learning Research. PMLR. https://proceedings.mlr.press/v202/li23q.html.

Li, Junnan, Dongxu Li, Caiming Xiong, and Steven Hoi. 2022. "BLIP: Bootstrapping Language-Image Pre-Training for Unified Vision-Language Understanding and Generation." In *Proceedings of the 39th International Conference on Machine Learning*, edited by Kamalika Chaudhuri, Stefanie Jegelka, Le Song, Csaba Szepesvari, Gang Niu, and Sivan Sabato, 162:12888–900. Proceedings of Machine Learning Research. PMLR. https://proceedings.mlr.press/v162/li22n.html.

Li, Zekun, Shiyang Li, and Xifeng Yan. 2023. "Time Series as Images: Vision Transformer for Irregularly Sampled Time Series." In *Advances in Neural Information Processing Systems*, edited by A. Oh, T. Naumann, A. Globerson, K. Saenko, M. Hardt, and S. Levine, 36:49187–204. Curran Associates, Inc. https://proceedings.neurips.cc/paper_files/paper/2023/file/9a17c1eb808cf012065e9db47b7ca80d-Paper-Conference.pdf.

Naveed, Humza, Asad Ullah Khan, Shi Qiu, Muhammad Saqib, Saeed Anwar, Muhammad Usman, Naveed Akhtar, Nick Barnes, and Ajmal Mian. 2024. "A Comprehensive Overview of Large Language Models." arXiv. https://doi.org/10.48550/arXiv.2307.06435.

Nepal, Subigya, Arvind Pillai, William Campbell, Talie Massachi, Michael V. Heinz, Ashmita Kunwar, Eunsol Soul Choi, et al. 2024. "MindScape Study: Integrating LLM and Behavioral Sensing for Personalized AI-Driven Journaling Experiences." *Proceedings of the ACM on Interactive, Mobile, Wearable and Ubiquitous Technologies* 8 (4): 1–44. https://doi.org/10.1145/3699761.

Nguyen, Viet Cuong, Mohammad Taher, Dongwan Hong, Vinicius Konkolics Possobom, Vibha Thirunellayi Gopalakrishnan, Ekta Raj, Zihang Li, et al. 2025. "Do Large Language Models Align with Core Mental Health Counseling Competencies?" In *Findings of the Association for Computational Linguistics: NAACL 2025*, edited by Luis Chiruzzo, Alan Ritter, and Lu Wang, 7488–7511. Albuquerque, New Mexico: Association for Computational Linguistics. https://aclanthology.org/2025.findings-naacl.418/.

"NHANES Homepage." 2024. https://www.cdc.gov/nchs/nhanes/index.htm.

Patterson, Matthew R., Adonay A. S. Nunes, Dawid Gerstel, Rakesh Pilkar, Tyler Guthrie, Ali Neishabouri, and Christine C. Guo. 2023. "40 Years of Actigraphy in Sleep Medicine and Current State of the Art Algorithms." *Npj Digital Medicine* 6 (1): 51. https://doi.org/10.1038/s41746-023-00802-1.

Pillai, Arvind, Dimitris Spathis, Subigya Nepal, Amanda C. Collins, Daniel M. Mackin, Michael V. Heinz, Tess Z. Griffin, Nicholas C. Jacobson, and Andrew Campbell. 2025. "Time2Lang: Bridging Time-Series Foundation Models and Large Language Models for Health Sensing Beyond Prompting." arXiv. https://doi.org/10.48550/arXiv.2502.07608.

Rahman, Syed Ashiqur, and Donald A. Adjeroh. 2019. "Deep Learning Using Convolutional LSTM

Estimates Biological Age from Physical Activity." *Scientific Reports* 9 (1): 11425. https://doi.org/10.1038/s41598-019-46850-0.

Ruan, Franklin, Stephen Adjei, Adaobi Amanna, George Price, Michael V. Heinz, and Nicholas C. Jacobson. 2024. "Characterizing Benzodiazepine Use in a Large National Study via Wearables and Deep Learning." PsyArXiv. https://doi.org/10.31234/osf.io/5ckme.

Ruan, Franklin Y., Aiwei Zhang, Jenny Y. Oh, SouYoung Jin, and Nicholas C. Jacobson. 2024. "AI Foundation Models for Wearable Movement Data in Mental Health Research." arXiv. https://doi.org/10.48550/ARXIV.2411.15240.

Sharma, Ashish, Inna W. Lin, Adam S. Miner, David C. Atkins, and Tim Althoff. 2023. "Human–AI Collaboration Enables More Empathic Conversations in Text-Based Peer-to-Peer Mental Health Support." *Nature Machine Intelligence* 5 (1): 46–57. https://doi.org/10.1038/s42256-022-00593-2.

Singhal, Karan, Shekoofeh Azizi, Tao Tu, S. Sara Mahdavi, Jason Wei, Hyung Won Chung, Nathan Scales, et al. 2023. "Large Language Models Encode Clinical Knowledge." *Nature* 620 (7972): 172–80. https://doi.org/10.1038/s41586-023-06291-2.

Xie, Qianqian, Qingyu Chen, Aokun Chen, Cheng Peng, Yan Hu, Fongci Lin, Xueqing Peng, et al. n.d. "Me-LLaMA: Foundation Large Language Models for Medical Applications." PhysioNet. Accessed June 2, 2025. https://doi.org/10.13026/WWFD-2T39.

Yoo, Jae-Young, Seyong Oh, Wissam Shalish, Woo-Youl Maeng, Emily Cerier, Emily Jeanne, Myung-Kun Chung, et al. 2023. "Wireless Broadband Acousto-Mechanical Sensing System for Continuous Physiological Monitoring." *Nature Medicine* 29 (12): 3137–48. https://doi.org/10.1038/s41591-023-02637-5.

# Appendix

## Appendix A: Label Generation And Evaluation
**Appendix A.1: Label Generation Prompt Template**
For generating MotionTeller Dataset labels, the following template was used given a sequence of 24 hourly data, aggravated from the original sequence of 1,440 raw values.

---

For this participant, we have 24 hours of hourly activity data, recorded as integer values. Each number represents the movement level during that specific hour of the day.

### **Participant-Specific Data:**
The participant's activity data begins in the first hour of the day, from midnight to 1AM.
- From 0000 to 0100, the activity level is {activity level 1}.
- From 0100 to 0200, the activity level is {activity level 2}.
- From 0200 to 0300, the activity level is {activity level 3}.
- From 0300 to 0400, the activity level is {activity level 4}.
- From 0400 to 0500, the activity level is {activity level 5}.
- From 0500 to 0600, the activity level is {activity level 6}.
- From 0600 to 0700, the activity level is {activity level 7}.
- From 0700 to 0800, the activity level is {activity level 8}.
- From 0800 to 0900, the activity level is {activity level 9}.
- From 0900 to 1000, the activity level is {activity level 10}.
- From 1000 to 1100, the activity level is {activity level 11}.
- From 1100 to 1200, the activity level is {activity level 12}.
- From 1200 to 1300, the activity level is {activity level 13}.
- From 1300 to 1400, the activity level is {activity level 14}.
- From 1400 to 1500, the activity level is {activity level 15}.
- From 1500 to 1600, the activity level is {activity level 16}.
- From 1600 to 1700, the activity level is {activity level 17}.
- From 1700 to 1800, the activity level is {activity level 18}.
- From 1800 to 1900, the activity level is {activity level 19}.
- From 1900 to 2000, the activity level is {activity level 20}.
- From 2000 to 2100, the activity level is {activity level 21}.
- From 2100 to 2200, the activity level is {activity level 22}.
- From 2200 to 2300, the activity level is {activity level 23}.
- From 2300 to 2400 which is 0000 of the next day, the activity level is {activity level 24}.

### **Global Context Awareness:**
- Across all participants within this participant's cohort, activity levels are scaled between 0 and 1000.
- This provides a reference range to interpret activity levels, but trends should still be analyzed based on this participant's specific data.
- If a participant's activity levels are all within a narrow range (for example, 0 to 20), describe this as low movement rather than assuming a buildup.

I would like you to offer an analysis on this participant's activity level throughout the 24 hour period. To ensure accuracy, use these steps to assist your reasoning process and provide a well-ordered,

consistent response:
1. First, count the number of 0s in the participants' activity data. If there are more than 16 activity levels that are 0, it is likely that the participant has misused their tracking device, or have forgotten to put on the device.
2. Next, start with a chronological breakdown of activity
    a. From the night to dawn (0000 to 0600)
    b. Early morning to noon (0600 - 1200)
    c. Noon through the entire afternoon (1200 - 1800)
    d. Early evening to midnight (1800 - 0000)
3. For each time period above, describe the general activity pattern by splitting the day into logical time periods. Keep descriptions qualitative and natural, without specifying exact hours or exact activity values
    a. Describe if the participant is mostly inactive, suggesting rest or sleep
    b. If activity is present, note whether it is sporadic or continuous.
    c. Mention when activity begins and whether it gradually increases or remains low.
    d. Describe the pattern of engagement, such as whether movement appears consistent or irregular.
    e. If movement increases, note whether the rise is steady or abrupt.
    f. If the activity is sustained, suggest a period of prolonged engagement.
4. Identify peak activity
    a. Identify when the highest level of movement occurs (without mentioning exact times). Note the general time block where engagement is most consistent.
5. Identify notable rest periods
    a. Highlight times of minimal or no activity, linking them to reasonable rest periods.
    b. If inactivity aligns with expected rest periods, acknowledge it.
6. Final summary statement
    a. Summarize the overall trend and what it suggests about the participant's routine.
    b. Avoid assumptions about specific activities.

Some additional rules to follow throughout generating the response:
- Use descriptive, narrative language, without formatting or lists
- Avoid exact values and time references
- Avoid jargon and technical language
- Do not make biased or overly strong statements

## Appendix A.2: Label Quality Evaluation Metrics

The following set of evaluation metrics are used in determining the quality of MotionTeller actigraphy description label output.

### Category 1: Identification of peak activity

Does the label correctly identify the period of highest movement?

| Score | Description |
| --- | --- |
| 5 | Perfect identification of the highest peak, correct time frame, and no mislabeling. |
| 4 | Identifies a local peak or the second-highest peak correctly, even if the main peak is missed; |

| | OR |
| --- | --- |
| | Peak identification is very close to the actual time, or some inconsistency shown; e.g., peak activity observed at 9 am, but mentioned both "early morning" or "mid morning" |
| 3 | Mentions the peak but lacks emphasis OR is ambiguous about its significance. |
| 2 | Misidentifies a different time period as the peak but still acknowledges fluctuations in activity. |
| 1 | Completely incorrect identification of peak OR no mention of peak activity at all. |

**Category 2: Description of nighttime to dawn (0000 - 0600)**
Does the label accurately capture the trends in activity levels during this early period?

| Score | Description |
| --- | --- |
| 5 | Clearly identifies the period, properly describes trends such as inactivity for example, and avoids unnecessary speculation |
| 4 | Mostly accurate, but slightly vague or missing minor details. |
| 3 | Overall trends, such as inactivity is acknowledged, but lacks depth OR contain minor inaccuracies |
| 2 | A mixture of right and wrong identifications of activity levels. |
| 1 | Completely wrong description OR no mention of this period. |

**Category 3: Description of early morning to noon (0600 - 1200)**
Does the label capture the gradual transition into wakefulness and engagement?

| Score | Description |
| --- | --- |
| 5 | Describes morning trends correctly, identifying gradual wakefulness or periods of movement appropriately. |
| 4 | Mostly accurate, but with minor wording or emphasis issues. |
| 3 | The description is somewhat vague OR lacks depth. |
| 2 | Misinterpretation of the morning pattern, such as exaggerating activity. |
| 1 | Completely incorrect description OR missing this section entirely. |

**Category 4: Description of afternoon activity (1200-1800)**
Does the label describe engagement levels accurately?

| Score | Description |
| --- | --- |
| 5 | Captures the afternoon's structure well, accurately highlighting patterns of movement and activity levels. |
| 4 | Mostly accurate, but with minor wording or emphasis issues. |
| 3 | The description is somewhat vague OR lacks depth. |
| 2 | Misinterpretation of the morning pattern, such as exaggerating activity. |
| 1 | Completely incorrect description OR missing this section entirely. |

**Category 5: Description of evening activity and end-of-day description (1800 - 0000)**
Does the label describe how activity patterns are during the night?

| Score | Description |
| --- | --- |
| 5 | Clearly describes how activity levels are during the night, avoiding unnecessary assumptions. |
| 4 | Mostly accurate, but with minor wording or emphasis issues. |
| 3 | The description is somewhat vague OR lacks depth. |
| 2 | Misinterpretation of the morning pattern, such as exaggerating activity. |
| 1 | Completely incorrect description OR missing this section entirely. |

**Category 6: Quality of language & bias avoidance**
Is the label written with clear, neutral, and descriptive language?

| Score | Description |
| --- | --- |
| 5 | The language is neutral, avoids bias, and maintains high clarity. |
| 4 | Mostly neutral and clear, but with some minor wording improvements needed. |
| 3 | Somewhat ambiguous wording or minor issues with neutrality. |
| 2 | Clear biases, unnecessary assumptions, or poor phrasing. |
| 1 | Contains misleading, incorrect, or highly biased language. |

## Appendix B: Understanding MIMS Units and Movement Intensity Thresholds

The National Health and Nutrition Examination Survey (NHANES) has established specific thresholds for classifying physical activity intensity levels using ActiGraph accelerometers. These thresholds are based on counts per minute (cpm) and are widely utilized in research to categorize activity levels.

NHANES ActiGraph cut points are generally defined as follows (Evenson and Wen 2015):

- Sedentary: <100 cpm
- Light Physical Activity: 100–2,019 cpm
- Moderate Physical Activity: 2,020–5,998 cpm
- Vigorous Physical Activity: ≥5,999 cpm

It is important to note that these cut points are based on vertical axis counts. With the advent of triaxial accelerometers, alternative thresholds have been proposed. For instance, the following triaxial thresholds were suggested in a study (Chomistek et al. 2017):

- Light Physical Activity: 200 to 2,689 cpm
- Moderate Physical Activity: 2,690 to 6,166 cpm
- Vigorous Physical Activity: ≥ 6,167 cpm

When interpreting MIMS (Monitor-Independent Movement Summary) units, one must recognize that they are derived differently from traditional ActiGraph counts. While some studies have attempted to equate MIMS units with ActiGraph counts, direct comparisons should be made cautiously. When working with MIMS data, researchers should consider the differences in measurement and apply appropriate methods to interpret activity levels accurately.